\documentclass{ifacconf}

\usepackage{natbib} 

\usepackage{graphicx} 
\usepackage{amsmath}

\usepackage{xcolor,soul,framed} %,caption
\colorlet{shadecolor}{yellow}
\usepackage{graphicx}
\usepackage{amsfonts}
\usepackage{multirow}
\usepackage{placeins}
\usepackage{nccmath}
\usepackage{cuted}
\usepackage{float}
\usepackage{lipsum}
\usepackage{appendix}
\usepackage{array}
\usepackage{mdwmath}
\usepackage{mdwtab}
\usepackage{eqparbox}
\usepackage{multicol}
\usepackage{url}
\usepackage{color}
\usepackage{url}

\newtheorem{definition}{Definition}

\usepackage{natbib}  % required for bibliography
\usepackage{lipsum,multicol}
\usepackage{dblfloatfix}

\begin{document}

\begin{frontmatter}

\title{Cascaded Tightly-Coupled Observer Design for Single-Range-Aided Inertial Navigation} 
\thanks[footnoteinfo]{This work was supported by the National Sciences and Engineering Research Council of Canada (NSERC), under the grants NSERC-DG RGPIN 2020-06270 and NSERC-DG RGPIN-2020-04759}
\author[First]{Oussama Sifour} 
\author[First,Third]{Soulaimane Berkane} 
\author[Third]{Abdelhamid Tayebi} 

\address[First]{Department of Computer Science
and Engineering, University of Quebec in Outaouais, Gatineau, QC, Canada. (e-mail: sifo01@uqo.ca, soulaimane.berkane@uqo.ca).}
%\address[Second]{Department of Computer Science and Engineering, University of Quebec in Outaouais, Gatineau, QC, Canada. (e-mail: )}
\address[Third]{Department of Electrical and Computer Engineering,  Lakehead University, Thunder Bay, ON P7B 5E1, Canada. (e-mail: atayebi@lakeheadu.ca)}

\begin{abstract}

This work introduces a single-range-aided navigation observer that reconstructs the full state of a rigid body using only an Inertial Measurement Unit (IMU), a body-frame vector measurement (\textit{e.g.,} magnetometer), and a distance measurement from a fixed anchor point. The design first formulates an extended linear time-varying (LTV) system to estimate body-frame position, body-frame velocity, and the gravity direction. The recovered gravity direction, combined with the body-frame vector measurement, is then used to reconstruct the full orientation on $\mathrm{SO}(3)$, resulting in a cascaded observer architecture. Almost Global Asymptotic Stability (AGAS) of the cascaded design is established under a uniform observability condition, ensuring robustness to sensor noise and trajectory variations. Simulation studies on three-dimensional trajectories demonstrate accurate estimation of position, velocity, and orientation, highlighting single-range aiding as a lightweight and effective modality for autonomous navigation.
\end{abstract}
\end{frontmatter}

\section{Introduction}

\subsection{Motivation and prior work}
Accurate estimation of  position, velocity, and attitude is critical for safe operation and reliable performance of Unmanned Aerial Vehicles (UAVs). State estimation forms the foundation of navigation, control, obstacle avoidance, and higher level autonomy and allows systems to effectively perform in uncertain or dynamical environments \citep{Titterton2004}. However, constructing observers for rigid body dynamics is still a challenging task because of the nonlinear coupling, sensors imperfections, and external disturbance typically presenting in practical situations.

 Industry standard estimation approaches typically rely on linearization and treat translational and rotational motions as two separate subsystems. For instance, \citep{Sabatini2006} linearized the attitude dynamics and estimated the rotation using a quaternion based EKF, which simplifies the rotational equations to reduce computational complexity. Building on such simplified models, \citep{Farrell2008, Whittaker2017} estimated the full state of the vehicle by processing translation and rotation independently, resulting in loosely coupled estimators. Although widely adopted due to their ease of implementation, loosely coupled approaches inherently assume weak interaction between  the translational and rotational dynamics. This assumption often fails during aggressive maneuvers, rapid attitude changes, or in the presence of significant disturbances, causing drift, inconsistency, and degraded robustness. 
 More recent efforts have been devoted to the design of tightly coupled full state observer designs, where position, velocity, and attitude are estimated simultaneously within a unified nonlinear tightly-coupled framework \citep{Johansen2018, BerkaneTayebi2019, Berkane2021}. By explicitly capturing the coupling between translational and rotational dynamics, these observers achieve significantly better robustness and accuracy than loosely coupled approaches.

To support inertial navigation, various sensing types have been explored. Global Navigation Satellite System (GNSS) provides absolute positioning outdoors \citep{Farrell2008, Grip2013, BerkaneTayebi2019}, but its performance deteriorates in cluttered, indoor, or urban environments due to blockage and multipath effects \citep{Misrabook2006}. Vision-based systems \citep{LiMourikis2013, Qin2018, Wang2022}  can deliver high-precision estimates but tend to be computationally heavy and sensitive to visual degradation such as motion blur or poor lighting. Range measurements, particularly Ultra-Wideband (UWB), offer an attractive low cost alternative by providing range measurements to fixed anchors \citep{Gryte2017, Hamer2018}. However, most UWB localization systems require multiple anchors to achieve 3D observability, which complicates the deployment. \citep{Mueller2015} showed that a  minimum of four non-coplanar anchors is typically needed to reconstruct a full three-dimensional position. 

Motivated by the desire to reduce sensing infrastructure, several works have investigated the minimal-sensing scenario in which the vehicle receives only a single distance measurement to a known point. The work in \citep{Hamel2017} deals with range-only localization using a nonlinear observer, where it is shown that a single range can be sufficient for position reconstruction when the motion is sufficiently informative. The work in \citep{Theodosis2021} considered a vehicle moving in three-dimensional space and proposed a state estimator design using one distance measurement. However, the orientation of the vehicle is not considered in their observer design.
In \citep{sifo2022}, a nonlinear observer capable of reconstructing the full rigid body state from a single body-frame bearing measurement was proposed. More recently, \citep{Wang2022} developed a vision-aided inertial navigation observer that operates with minimal visual information, showing that low dimensional optical cues can be sufficient for full state recovery when combined with inertial data. Collectively, these works establish that, in such highly constrained sensing scenarios, observability critically depends on a persistency of excitation (PE) condition, requiring the motion to be sufficiently rich over time for the state to be uniquely reconstructed.
\subsection{Contributions and organization}
This work develops a single-range-aided inertial navigation observer that reconstructs the full state of a rigid body in three-dimensional space. The key idea is a dynamic state augmentation that transforms the nonlinear range constraint into a linear
time-varying output. By introducing four auxiliary coordinates, the measured range and its time derivatives become linear functions of the augmented state while their dynamics remain driven only by measured quantities. This construction enables
the application of a Riccati observer framework, inspired by earlier work on range-based estimation \citep{Hamel2017,Theodosis2021}, to recover position, velocity, and the gravity direction in body
coordinates. The recovered gravity direction is then combined with an auxiliary body-frame vector measurement (\textit{e.g.,} magnetometer) in a complementary filter on $\mathrm{SO(3)}$, yielding a cascaded tightly coupled observer that reconstructs the full orientation. We establish uniform observability of the LTV subsystem under a persistency of excitation condition depending only on angular velocity and acceleration, and we prove Almost Global Asymptotic Stability of the overall cascaded design. Simulation results on representative trajectories illustrate the accuracy and robustness of the proposed approach in the presence of sensor noise. The remainder of this paper is organized as follows. In section \ref{section2}, we present the mathematical notations .Section \ref{section3}, we present the system model and problem formulation, including the vehicle’s dynamics and the measurement models. Section \ref{section4} details the observer design, describing the Riccati observer for position and velocity estimation and the attitude observer for orientation estimation. Section \ref{sectionUO}
establishes the uniform observability of the proposed observer. Section \ref{section6} shows simulation results

\section{Preliminaries and Notation} \label{section2}
We denote by $\mathbb{R}$ the set of real numbers and by $\mathbb{R}^n$ the $n$-dimensional Euclidean space. The Euclidean norm of a vector $x \in \mathbb{R}^n$ is denoted by $\|x\|$, and the $i$-th component of a vector $x$ is denoted by $x_i$. The identity matrix of size $n$ is written as $I_n$, and $0_{m \times n}$ denotes the $m \times n$ zero matrix. The notation $\text{blkdiag}(M_1, \dots, M_k)$ refers to the block diagonal matrix formed by matrices $M_1$ to $M_k$. The unit sphere in $\mathbb{R}^n$ is denoted by $\mathbb{S}^{n-1} := \{x \in \mathbb{R}^n : \|x\| = 1\}$. The kernel (null space) of a matrix $M$ is denoted $\ker(M) := \{x \in \mathbb{R}^n : Mx = 0\}$, and the image  of $M$ is denoted $\operatorname{Im}(M) := \{Mx : x \in \mathbb{R}^n\}$, the image of a subspace $\mathbb{L} \subset \mathbb{R}^n$ is denoted $\operatorname{Im}_{\mid \mathbb{L}}(M) := \{Mx : x \in \mathbb{L}\}$. The Special Orthogonal Group $\mathrm{SO}(3) := \{R \in \mathbb{R}^{3 \times 3} : R^\top R = I_3,\ \det(R) = 1\}$ represents the space of 3D rotation matrices. The associated Lie algebra is $\mathfrak{so}(3) := \{\Omega \in \mathbb{R}^{3 \times 3} \mid \Omega = -\Omega^\top\}$. For any $x \in \mathbb{R}^3$, the skew-symmetric matrix $[x]_\times$ is defined such that $[x]_\times y = x \times y$ for all $y \in \mathbb{R}^3$, where $\times$ denotes the cross product.  The inverse isomorphism of the map $[\cdot]_{\times}$is defined by $\mathrm{vex}: \mathfrak{s o}(3) \rightarrow \mathbb{R}^3$, such that $\mathrm{vex}\left([\omega]_{\times}\right)=\omega$, for all $\omega \in \mathbb{R}^3$ and $[\mathrm{vex}(\Omega)]_{\times}=\Omega$, for all $\Omega \in \mathfrak{s o}(3)$. The composition map $\psi_a:=\mathrm{vex} \circ \mathbf{P}_{\mathfrak{s} \mathfrak{o}(3)}$ extends the definition of $\mathrm{vex}$ to $\mathbb{R}^{3 \times 3}$, where $\mathbf{P}_{\mathfrak{s} \mathfrak{o}(3)}: \mathbb{R}^{3 \times 3} \rightarrow \mathfrak{s o}(3)$ is the projection map on the Lie algebra $\mathfrak{s o}(3)$ such that $\mathbf{P}_{\mathfrak{s o}(3)}(A):=\left(A-A^{\top}\right) / 2$. Accordingly, for a 3-by-3 matrix $A:=\left[a_{i j}\right]_{i, j=1,2,3}$, one has $\psi_a(A):=\mathrm{vex}\left(\mathbf{P}_{\mathfrak{s o}(3)}(A)\right)=\frac{1}{2}\left[a_{32}-a_{23}, a_{13}-a_{31}, a_{21}-\right. \left.a_{12}\right]$. We define the standard basis vectors of $\mathbb{R}^3$ as $e_1 := [1\ 0\ 0]^\top$, $e_2 := [0\ 1\ 0]^\top$, and $e_3 := [0\ 0\ 1]^\top$. The projection matrix onto the plane orthogonal to a unit vector $x \in \mathbb{S}^2$ is given by $\Pi_x := I_3 - x x^\top$.

\section{Problem Formulation} \label{section3}
We consider a rigid body (vehicle) to be moving freely in a three-dimensional (3D) space. We denote $\{\mathcal{I}\}$ as an inertial reference frame and $\{\mathcal{B}\}$ as a body-fixed frame attached to the center of mass of the vehicle. The orientation (attitude) of the body frame with respect to the inertial frame is given by the rotation matrix $R \in \mathrm{SO}(3)$. The kinematic equations that govern the motion of the rigid body are:
\begin{subequations}
\begin{align}
\dot{{p}}^{\mathcal{I}} &= {v}^{\mathcal{I}}, \label{eq:position_kinematics} \\
\dot{{v}}^{\mathcal{I}} &= {g}^{\mathcal{I}} + R {a}^{\mathcal{B}}, \label{eq:velocity_kinematics} \\
\dot{R} &= R [\omega]_{\times}, \label{eq:rotation_kinematics}
\end{align}
\end{subequations}
where ${p}^{\mathcal{I}} \in \mathbb{R}^3$ is the position vector of the rigid body expressed in the inertial frame $\{\mathcal{I}\}$, ${v}^{\mathcal{I}} \in \mathbb{R}^3$ is the linear velocity of the rigid body expressed in $\{\mathcal{I}\}$, ${g}^{\mathcal{I}} \in \mathbb{R}^3$ is the gravity vector expressed in $\{\mathcal{I}\}$, ${a}^{\mathcal{B}} \in \mathbb{R}^3$ is the apparent acceleration (non-gravitational forces) expressed in the body frame $\{\mathcal{B}\}$, ${\omega} \in \mathbb{R}^3$ is the angular velocity of the body frame $\{\mathcal{B}\}$ with respect to the inertial frame $\{\mathcal{I}\}$, expressed in $\{\mathcal{B}\}$. 
Equation \eqref{eq:position_kinematics} describes the evolution of the position based on the velocity. Equation \eqref{eq:velocity_kinematics} accounts for gravitational acceleration and apparent acceleration due to control inputs or disturbances. Equation \eqref{eq:rotation_kinematics} captures the rotational dynamics of the vehicle.

We assume the rigid body is equipped with an IMU, a single UWB range receiver, 
and a body-frame vector sensor (\textit{e.g.,} a magnetometer). 
The IMU provides angular velocity and apparent acceleration measurements, 
while the vector sensor provides directional information with respect to a known inertial reference. 
The corresponding measurement models are  
\begin{equation}
\begin{aligned}
\omega_y &= \omega + \eta_\omega, \\
a_y^{\mathcal{B}} &= R^\top a^{\mathcal{I}} + \eta_{\text{acc}}, \\
m_y^{\mathcal{B}} &= R^\top m^{\mathcal{I}} + \eta_{\text{mag}},
\end{aligned}
\end{equation}
where $\omega_y$, $a_y^{\mathcal{B}}$, and $m_y^{\mathcal{B}}$ denote the measured angular velocity, 
apparent acceleration, and body-frame vector, respectively. 
The terms $\eta_\omega$, $\eta_{\text{acc}}$, and $\eta_{\text{mag}}$ represent zero-mean Gaussian white noises. 
The vector $a^{\mathcal{I}}$ is the apparent acceleration in the inertial frame, 
while $m^{\mathcal{I}}$ is a known constant inertial vector (\textit{e.g.,} the Earth's magnetic field).  

\begin{figure}[h]
    \centering
    \includegraphics[width=0.85\linewidth]{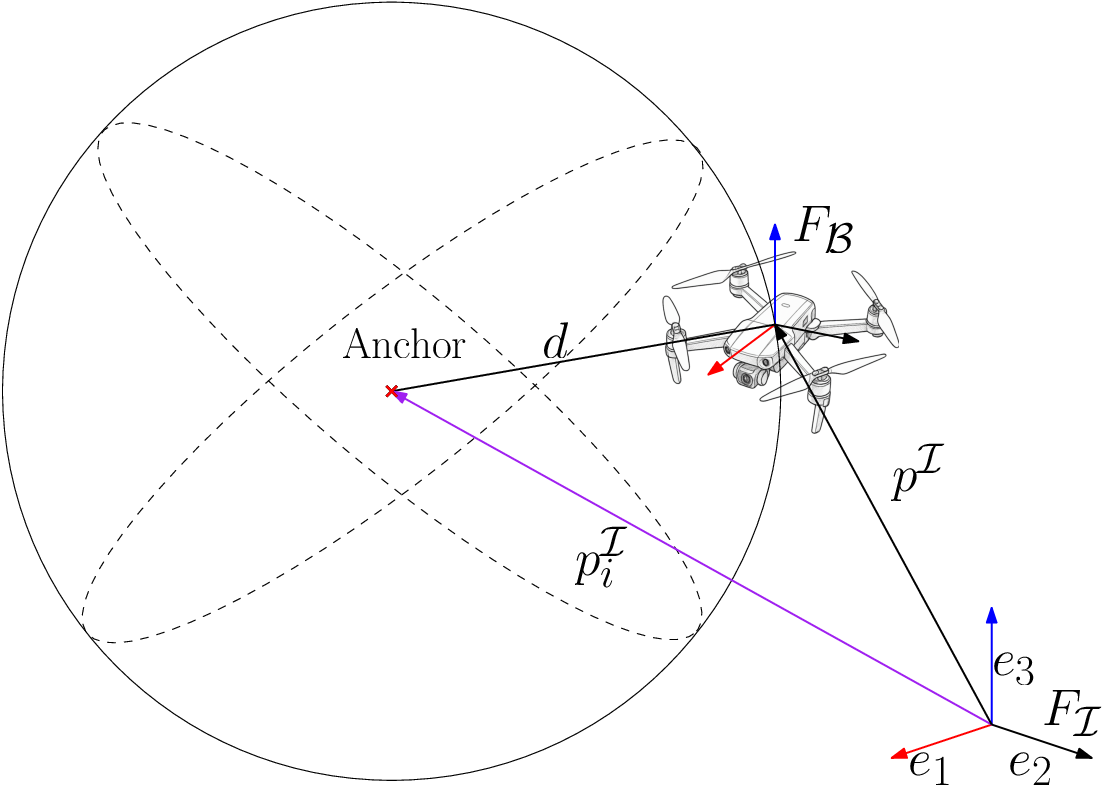}
\caption{Illustration of a single range measurement situation.}
    \label{figsingle}
\end{figure}

The UWB receiver provides a scalar range measurement to a fixed anchor located at $p_i^{\mathcal{I}}$ as illustrated in figure \ref{figsingle}, the range measurement is given by
\begin{equation}
d_y = \| p^{\mathcal{I}} - p_i^{\mathcal{I}} \| + \eta_{\text{uwb}},
\end{equation}
with $\eta_{\text{uwb}}$ denoting the measurement noise. 
Each range defines a sphere of possible positions centered at the anchor, 
and thus a single range measurement alone does not uniquely determine the position.  

For the deterministic observer design that follows, stochastic noise is neglected 
($\eta_* \equiv 0$) to simplify the analysis without affecting the generality of the framework. 
The objective is to design a deterministic observer that reconstructs the full state of the rigid body---position, 
velocity, and orientation---using IMU data, a single range measurement, 
and one body-frame vector measurement. 
The proposed approach ensures uniform observability under suitable excitation conditions 
and guarantees Almost Global Asymptotic Stability of the resulting state estimates.

\section{Main results} \label{section4}
\begin{figure*}[!t]
    \centering
    \includegraphics[width=0.85\textwidth]{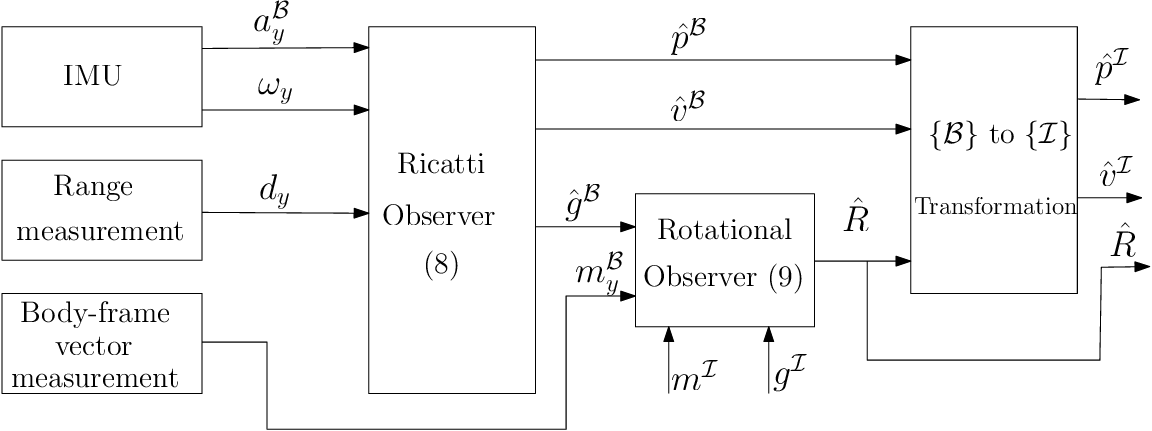}
    \caption{Illustration of the proposed state estimation approach.} 
    \label{fig:observer_illustration}
\end{figure*}

The key step in the proposed observer design is to recast the nonlinear estimation problem into a linear time-varying (LTV) framework where Riccati-type observers can be applied. To this end, we express the dynamics in the body frame and introduce a minimal dynamic augmentation of four scalar variables that lifts the nonlinear range constraint into a linear output. This augmentation preserves the system’s structure while enabling the incorporation of the range measurement in a linear form. Building on this representation, we construct a Riccati observer for position, velocity, and gravity direction, which is then cascaded with a nonlinear attitude observer on $\mathrm{SO}(3)$ to recover the full orientation, see Figure \ref{fig:observer_illustration}. The remainder of this section details the construction of the augmented model, the resulting LTV formulation, and the cascaded observer design together with its stability analysis.
\subsection{Augmented Body-Frame Model}
We first express the kinematic equations in the body frame $\{\mathcal{B}\}$ as
\begin{subequations}\label{eq:body_kinematics}
\begin{align}
\dot{p}^{\mathcal{B}} &= -[\omega]_{\times}\,p^{\mathcal{B}} + v^{\mathcal{B}}, \label{eq:body_position_dynamics}\\
\dot{v}^{\mathcal{B}} &= -[\omega]_{\times}\,v^{\mathcal{B}} + g^{\mathcal{B}} + a^{\mathcal{B}}, \label{eq:body_velocity_dynamics}\\
\dot{g}^{\mathcal{B}} &= -[\omega]_{\times}\,g^{\mathcal{B}}. \label{eq:body_gravity_dynamics}
\end{align}
\end{subequations}
Here, $p^{\mathcal{B}} = R^\top p^{\mathcal{I}}$, $v^{\mathcal{B}} = R^\top v^{\mathcal{I}}$, and $g^{\mathcal{B}} = R^\top g^{\mathcal{I}}$ denote the position, velocity, and gravity vectors expressed in the body frame. Now, let the composite state vector be defined as
$
\bar{x}^{\mathcal{B}} := \operatorname{col}\!\big(p^{\mathcal{B}},\, v^{\mathcal{B}},\, g^{\mathcal{B}}\big) \in \mathbb{R}^9.
$
Then, the system can be compactly expressed as an LTV system with a quadratic output:
\begin{equation}\label{eq:LTV_system}
\begin{aligned}
\dot{\bar{x}}^{\mathcal{B}} &= A(t)\,\bar{x}^{\mathcal{B}} + \bar{B}\,a^{\mathcal{B}}, \\[4pt]
y &= \tfrac{1}{2}\,(\bar{x}^{\mathcal{B}})^\top C\,\bar{x}^{\mathcal{B}},
\end{aligned}
\end{equation}
where $y := \tfrac{1}{2}d_y^2$ is the squared range output, and
\[
A(t) = \bar{A} - \operatorname{blkdiag}\!\big([\omega]_{\times},[\omega]_{\times},[\omega]_{\times}\big),
\quad
\bar{B} =
\begin{bmatrix}
0_{3\times3}\\[2pt]
I_3\\[2pt]
0_{3\times3}
\end{bmatrix},
\]
\[
\bar{A} =
\begin{bmatrix}
0_{3\times3} & I_3 & 0_{3\times3}\\
0_{3\times3} & 0_{3\times3} & I_3\\
0_{3\times3} & 0_{3\times3} & 0_{3\times3}
\end{bmatrix},
\qquad
C =
\begin{bmatrix}
I_3 & 0 & 0\\
0 & 0 & 0\\
0 & 0 & 0
\end{bmatrix}.
\]
To incorporate the nonlinear range measurement into this framework, we introduce a dynamic state augmentation that lifts the quadratic range constraint into a linear time-varying output. Following the approach of \citep{Theodosis2021}, we add four auxiliary coordinates that collect the algebraic terms arising when differentiating the range signal along the system kinematics. This minimal augmentation ensures that the measured range and its time derivatives become linear functions of the augmented state, while the corresponding dynamics remain driven only by measured quantities.
We introduce four auxiliary coordinates
\[
\xi_i = \tfrac{1}{2}\,\bar{x}^{\mathcal B\top} C_i \bar{x}^{\mathcal B}, 
\qquad i = 1,\dots,4,
\]
where $C_1 = C$ and $C_{i+1} = C_i A(t) + A(t)^\top C_i$ for $i=1,\dots,4$. 
Differentiating $\xi_i$ and using the recursive definition of $C_i$ yield
\begin{subequations}\label{eq:xi_dynamics}
\begin{align}
\dot{\xi}_i &= \tfrac{1}{2}\,\bar{x}^{\mathcal B\top}\!\big(A^\top C_i + C_i A\big)\bar{x}^{\mathcal B} 
               + a^{\mathcal B\top}\bar{B}^\top C_i \bar{x}^{\mathcal B} \\[2pt]
            &= \xi_{i+1} + a^{\mathcal B\top}\bar{B}^\top C_i \bar{x}^{\mathcal B}, 
            \quad i = 1,2,3, \\
\dot{\xi}_4 &= a^{\mathcal B\top}\bar{B}^\top C_4 \bar{x}^{\mathcal B},
\end{align}
\end{subequations}
where $C_5 \equiv 0$ has been used to simplify the last equation.  

Collecting all coordinates, we define the augmented state vector
\[
x^{\mathcal B} := \operatorname{col}(\xi_1,\xi_2,\xi_3,\xi_4,\bar{x}^{\mathcal B}) \in \mathbb{R}^{13}.
\]
The resulting augmented system can be expressed in linear time-varying form as
\begin{subequations}\label{aug_system}
\begin{align}
\dot{x}^{\mathcal B} &= \mathcal{A}(t)\,x^{\mathcal B} + \mathcal{B}\,u(t), \label{eq:augmented_system}\\[2pt]
y &= \mathcal{C}\,x^{\mathcal B} := [\,\mathcal{C}_m \;\; 0_{1\times9}\,]\,x^{\mathcal B}, \label{eq:output_equation}
\end{align}
\end{subequations}
where $u(t) = a^{\mathcal B}$, and the matrices 
$\mathcal{A}(t)$, $\mathcal{B}$, and $\mathcal{C}$ follow directly from~\eqref{eq:xi_dynamics} and~\eqref{aug_system}.

where the system matrices are given by
\[
\mathcal{A}(t)=
\begin{bmatrix}
\mathcal{S} & \mathcal{T}(t)\\[2pt]
0_{9\times4} & A(t)
\end{bmatrix},\qquad
\mathcal{B}=
\begin{bmatrix}
0_{4\times3} & B_m\\[2pt]
\bar{B} & 0_{9\times1}
\end{bmatrix},
\]
with
\[
\mathcal{S}=
\begin{bmatrix}
0&1&0&0\\
0&0&1&0\\
0&0&0&1\\
0&0&0&0
\end{bmatrix},\qquad
\mathcal{T}(t)=
\begin{bmatrix}
a^{\mathcal B\top}\bar{B}^\top C_1 \\
a^{\mathcal B\top}\bar{B}^\top C_2 \\
a^{\mathcal B\top}\bar{B}^\top C_3 \\
a^{\mathcal B\top}\bar{B}^\top C_4
\end{bmatrix}.
\]
Here the input vector is given by $u(t)=\operatorname{col}(a^{\mathcal B},\|g^{\mathcal B}\|^2)$,  
and the small blocks by $B_m=[\,0\;0\;0\;1\,]^\top$, $\mathcal{C}_m=[\,1\;0\;0\;0\,]$. The output $y$ consists of measurable quantities; in particular, its first component $\xi_1$ corresponds directly to the UWB range measurement.

\subsection{Proposed Observer Architecture}
We now design the complete cascaded observer, which combines a Riccati observer for the augmented translational LTV system with a complementary filter for attitude estimation on $\mathrm{SO}(3)$. This design provides full-state estimation (position, velocity, gravity, and orientation) from IMU and single-range measurements. The main stability result is then stated under a uniform observability (UO) assumption, which will be analyzed in the subsequent section. The Riccati observer for the augmented system \eqref{aug_system} is given by
\begin{equation}\label{eq:riccati}
\dot{\hat{x}}^{\mathcal B} = \mathcal{A}(t)\hat{x}^{\mathcal B} + \mathcal{B}u(t) + K(t)\left(y - \mathcal{C}\hat{x}^{\mathcal B}\right),
\end{equation}
where $\hat{x}^{\mathcal B}$ is the estimate of the augmented state vector and $K(t)=P(t)\mathcal{C}^\top Q(t)$ is the observer gain. The gain matrix $P(t)$ evolves according to the differential Riccati equation
\[
\dot{P}(t) = \mathcal{A}(t)P(t) + P(t)\mathcal{A}^\top(t) - P(t)\mathcal{C}^\top Q(t)\mathcal{C} P(t) + V(t),
\]
with $P(0)>0$ and $Q(t),V(t)$ uniformly positive definite. This observer provides estimates of position, velocity, and the gravity vector expressed in the body frame.

The estimated gravity vector is then combined with the magnetometer measurement in a complementary filter to reconstruct the attitude on $\mathrm{SO}(3)$ (see \cite{Mahony2008}):
\begin{subequations}
\begin{align}\label{eq:att}
\dot{\hat{R}} &= \hat{R}[\omega_y+k_1\sigma]_{\times},\\
\sigma &=\rho_1\left(m^{\mathcal B}_y \times \hat{R}^\top m^{\mathcal I}\right) + \rho_2\left(\hat{g}^{\mathcal B} \times \hat{R}^\top g^{\mathcal I}\right),
\end{align}
\end{subequations}
where $\omega_y$ is the measured angular velocity, $m^{\mathcal B}_y$ is the magnetometer measurement, $\hat{g}^{\mathcal B}$ is the gravity estimate from \eqref{eq:riccati}, and $k_1,\rho_1,\rho_2$ are positive tuning parameters. This filter mitigates orientation drift by realigning the estimated attitude
with the measured inertial directions. The resulting closed-loop error dynamics are
\begin{subequations}\label{closedloop}
\begin{align}
\dot{\tilde x}^{\mathcal B} &= \left(\mathcal A(t) - K(t)\,\mathcal C\right)\,\tilde x^{\mathcal B}, \label{closedloop_x}\\[1mm]
\dot{\tilde R} &= \tilde R
\left( -k_1\,\hat R\,\psi_a\!\left(M_\pi \tilde R\right)
   + \Gamma(t)\,\tilde x^{\mathcal B}
\right)_{\times}, \label{closedloop_R}
\end{align}
\end{subequations}
where $\tilde R := R \hat R^\top \in \mathrm{SO}(3)$ is the right-invariant attitude error and
$\tilde x^{\mathcal B} := x^{\mathcal B} - \hat x^{\mathcal B}$ is the estimation error of the augmented state. Here $M_\pi := \rho_1\, m^{\mathcal I} m^{\mathcal I\top}
+ \rho_2\, g^{\mathcal I} g^{\mathcal I\top}$ and
$\Gamma(t) := -k_1 \rho_2\, \hat R \,[\tilde R^\top g^{\mathcal B}]_{\times} L_g$
which involves the linear map $L_g$ defined by $\hat g^{\mathcal B} = L_g \hat x^{\mathcal B}$.
The following theorem characterizes the convergence properties of the closed-loop system.
\begin{thm}
Assume that $(\mathcal{A}(t),\mathcal{C})$ is uniformly observable (see Section~\ref{sectionUO}), and that the inertial vectors $m^{\mathcal I}$ and $g^{\mathcal I}$ are known, constant, and non-collinear. Then the estimation errors of the cascaded observer \eqref{eq:riccati}–\eqref{eq:att} satisfy:
\begin{itemize}
    \item [\textit{i})] The estimation error $(\tilde{R},\tilde {x}^{\mathcal B})$ converges to the set $\left(\{I_3\}\cup\mathcal{U}_{\pi}\right)\times\{0\}$, where $\mathcal{U}_{\pi} := \{\,I_3-2uu^\top \mid u\in\mathbb S^2 \text{ and } u \text{ is an eigenvector of } M_{\pi}\,\}$ is the set of $\pi$-rotations about eigen-directions of $M_{\pi}$.
    \item[\textit{ii})] The desired equilibrium $(\tilde R,\tilde x^{\mathcal B})=(I_3,0)$ is locally exponentially stable.
    \item[\textit{iii})] All equilibria in $\mathcal U_\pi$ are unstable and the desired equilibrium is almost globally asymptotically stable.
\end{itemize}
\end{thm}

\begin{pf}[Sketch]
The closed-loop error dynamics of the proposed cascaded observer have the same structure as those analyzed in \cite{Wang2022}, consisting of (i) a GES translational subsystem driven by the Riccati-based observer, and (ii) an AGAS rotational subsystem given by a complementary filter on $\mathrm{SO}(3)$. Under uniform observability, the translational estimation error converges exponentially, and the resulting rotational error dynamics reduce asymptotically to the same nonlinear form studied in \cite{Wang2022}, with the same set of unstable $\pi$-rotation equilibria.

Since the two subsystems are interconnected in cascade and the translational subsystem is GES, the perturbation acting on the rotational dynamics vanishes asymptotically. By standard cascaded stability arguments (see Appendix~A of \cite{Wang2022}), AGAS of the full estimation error follows. The detailed proof is omitted for brevity.
\end{pf}

\section{Observability Analysis} \label{sectionUO}
To ensure accurate state estimation, it is essential that the augmented system be uniformly observable. This section establishes sufficient conditions for uniform observability by taking advantage of the structure of the system and simplifying the associated analysis.

\begin{definition}[Uniform Observability] \label{def:uniform_obs}
The system \eqref{aug_system} is said to be \textit{uniformly observable} if there exist constants $\delta, \mu > 0$ such that, for all $t \geq 0$:  
\begin{equation}  
W(t, t + \delta) \geq \mu I_{13} \geq 0, 
\label{eq:main Uniform Observability condition}
\end{equation}  
where  
\begin{equation} \label{eq:observability_gramian}
W(t, t + \delta) := \frac{1}{\delta} \int_{t}^{t+\delta} \Phi^\top(s, t)\mathcal{C}^\top(s)\mathcal{C}(s)\Phi(s, t) \, ds %\notag 
\end{equation} is the observability Gramian, and  
$\Phi(s, t)$ denotes the state transition matrix associated with $\mathcal{A}(t)$, defined by: $\frac{d}{dt} \Phi(s, t) = \mathcal{A}(t) \Phi(s, t)$, $\Phi(t, t) = I_{13}$.
If condition \eqref{eq:main Uniform Observability condition} holds, the pair $(\mathcal{A}(t), \mathcal{C}(t))$ is said to be \textit{uniformly observable}.
\end{definition} Since the matrix $A(t)$ has an upper triangular block, the state transition $\Phi(t, s)$ follows the same structure and is given by 
\[
\Phi(\tau, t) = \begin{bmatrix}
\phi_{11}(\tau, t) & \phi_{12}(\tau, t) \\
0 & \phi_{22}(\tau, t)
\end{bmatrix},
\]
where \(\phi_{11} \in \mathbb{R}^{4 \times 4}\), \(\phi_{12} \in \mathbb{R}^{4 \times 9}\), and \(\phi_{22} \in \mathbb{R}^{9 \times 9}\). According to \citep{Theodosis2021}, these components satisfy the differential equations:
\begin{align*}
\frac{d \phi_{11}(\tau, t)}{d \tau} &= \mathcal{S} \phi_{11}(\tau, t), \quad \phi_{11}(\tau, \tau) = I_4, \\
\frac{d \phi_{12}(\tau, t)}{d \tau} &= \mathcal{S} \phi_{12}(\tau, t) + \mathcal{T}(\tau) \phi_{22}(\tau, t), \quad \phi_{12}(\tau, \tau) = 0, \\
\frac{d \phi_{22}(\tau, t)}{d \tau} &= A(t)(\tau) \phi_{22}(\tau, t), \quad \phi_{22}(\tau, \tau) = I_{9}.
\end{align*}
To facilitate the analysis of uniform observability, we reformulate the problem by reducing the observability of the augmented system to that of a simpler, lower-dimensional system. This transformation preserves the essential observability properties and significantly simplifies the analytical complexity. The equivalence is formalized in the following proposition.
\begin{prop} \label{prop1}
Let $r_0 = 0$ and 
\[
\begin{aligned}
r_{i+1}(t) &= r_i(t)\,A(t) + \dot{r}_i(t)
              + \mathcal{C}_m \mathcal{S}^i \mathcal{T}(t),
\quad i = 0,1,\dots,4 .
\end{aligned}
\]
Assume that the apparent acceleration and angular velocity are bounded and differentiable, and that the inertial vectors $m^{\mathcal I}$ and $g^{\mathcal I}$ are known, constant, and non-collinear.  
Then the system~\eqref{aug_system} is uniformly observable if the reduced pair $(\bar A,\; \bar R\, r_4)$ is uniformly observable, where $\bar R$ is defined as \(\bar{R} := \mathrm{blkdiag}(R, R, R)\). 
\end{prop}
The proof of this proposition is provided in Appendix~\ref{appendixA}. 
This result is sufficient to reduce the observability analysis to verifying the uniform observability of the lower-dimensional pair $(\bar{A},\, \bar{R} r_4)$, rather than the full time-varying augmented system $(\mathcal{A}(t),\, \mathcal{C})$. 
Nevertheless, this reduction still requires evaluating the associated observability Gramian. 
Our next objective is to establish a persistence-of-excitation (PE) condition that guarantees uniform observability \emph{without} explicitly computing this Gramian. 
To this end, we introduce the following technical lemma, which is itself a contribution of this work.
\begin{lem}\label{TechnicalLemma}
Let $A \in \mathbb{R}^{n\times n}$, $H \in \mathbb{R}^{m\times n}$, and $\Theta : \mathbb{R}_{\ge 0} \to \mathbb{R}^{m\times m}$ such that:
\begin{itemize}
\item $(A,H)$ is Kalman observable
\item $A$ is nilpotent, \textit{i.e.,} $A^q = 0$ for some $q \le n$
\item $\Theta(t)$ is uniformly bounded for all $t \ge 0$
\end{itemize}
Then, the pair $(A,\Theta(t)H)$ is uniformly observable if there exist $\delta, \mu > 0$ such that
\[
\forall t : \ \frac{1}{\delta} \int_{t}^{t+\delta} \|\Theta(s)z\|^2 \, ds \ \ge \ \mu,\quad \forall z \in \mathbb{E} \cap \mathbb{S}^{m-1},
\]
where $\mathbb{E} \subseteq \mathbb{R}^m$ is the set defined by
\[
\mathbb{E} \ :=\ \bigcup_{k=0}^{q-1} \operatorname{Im}_{\mid L_k}(H A^{k}), 
\qquad 
L_k \ =\ \bigcap_{i \ge k+1} \operatorname{Ker}(H A^i).
\]

\end{lem}
The proof of this technical lemma is provided in Appendix~\ref{appendixB}. 
We next invoke Lemma~\ref{TechnicalLemma} to establish an explicit PE condition for the reduced pair $(\bar{A},\, \bar{R}(t)\, r_4(t))$. 
By Proposition~\ref{prop1}, this PE condition is sufficient to guarantee the uniform observability of the original augmented system characterized by the pair $(\mathcal{A}(t),\, \mathcal{C})$.
\begin{lem}\label{lemma1}
The system given by \eqref{aug_system} is uniformly observable if there exist $\delta_5, \mu_5>0$ such that
\begin{align}
\int_{t}^{t+\delta_5}\phi(s)\phi(s)^\top ds \geq \mu_5 I_3
\end{align}
with $\phi(t)=\ddot a^{\mathcal B}(t)-[\omega(t)]_\times\dot a^{\mathcal B}(t)-2[\omega(t)]_\times^2 a^{\mathcal B}(t)\in\mathbb{R}^3$
\end{lem}
The proof of this Lemma is given in appendix \ref{appendixC}

The term $[\dot{\omega}]_{\times}$ captures the influence of rotational dynamics on the apparent acceleration, while $\ddot{a}^{\mathcal{B}}$ denotes the second derivative of the apparent acceleration. 
This sufficient condition indicates that the apparent acceleration $a^{\mathcal{B}}$ and its time derivatives must exhibit enough variation over the interval $[t_0,\, t_0+\delta]$ to ensure observability. In practice, satisfying this condition typically involves two aspects. 
First, appropriate trajectory design can help introduce motion that excites the relevant modes of the system. 
Second, control-input modulation—through adequately varying $a^{\mathcal{B}}$—can contribute the required level of excitation. 
Together, these considerations help ensure that the sufficient observability condition is met.

\section{Simulation} \label{section6}
To validate the performance of the proposed observer, we conduct simulations using a vehicle following a predefined trajectory and subject to realistic sensor noise. The vehicle moves in 3D space following an eight-shaped trajectory defined by: $$
{p}^{\mathcal{I}}(t) = \begin{bmatrix} \cos(8t), &
\frac{3}{10}\,\sin\!\left(16t + \frac{\pi}{12}\right), &
\frac{-\sqrt{3}}{4}\,\sin\!\left(16t - \frac{\pi}{9}\right)\end{bmatrix}^\top.$$ This trajectory provides sufficient excitation in all three spatial dimensions. The vehicle's angular velocity is given by
$
{\omega}(t) = \begin{bmatrix} \sin(0.1t + \pi),& 0.5 \sin(0.2t), & 0.1 \sin\left(0.3t + \dfrac{\pi}{3}\right) \end{bmatrix}^\top.
$
This angular velocity introduces time-varying rotation, contributing to the excitation needed for observability. For the true states, the position is initialized as ${p}^{\mathcal{I}}(0) = \begin{bmatrix} 1 & 0 & 0 \end{bmatrix}^\top$, the velocity as ${v}^{\mathcal{I}}(0) = \begin{bmatrix} -0.0125 & 2.5 & -4.33 \end{bmatrix}^\top$, and the orientation as $R(0) = \exp\left( \dfrac{\pi}{2} {[e_2]}_{\times} \right)$. For the observer estimates, the initial estimated states are $\hat{{x}}(0) = 0_{13 \times 1}$, and the initial  orientation condition is $\hat{R}(0) = I_{3}$. The gravity vector is set as ${g}^{\mathcal{I}} = \begin{bmatrix} 0 & 0 & 9.81 \end{bmatrix}^\top \, \text{m/s}^2$, and the magnetic field as ${m}^{\mathcal{I}} = \dfrac{1}{\sqrt{2}} \begin{bmatrix} 1 & 0 & 1 \end{bmatrix}^\top$. The sensor noises are modeled as zero-mean Gaussian white noise. The gyroscope noise has variance $\sigma_\omega^2$, the accelerometer noise has variance $\sigma_{\text{acc}}^2$, the magnetometer noise has variance $\sigma_{\text{mag}}^2 = 10^{-2}$, and the UWB noise has variance $\sigma_{\text{uwb}}^2$.

\begin{figure}%[h]
    \centering
    \includegraphics[width=1\columnwidth,height=6cm]{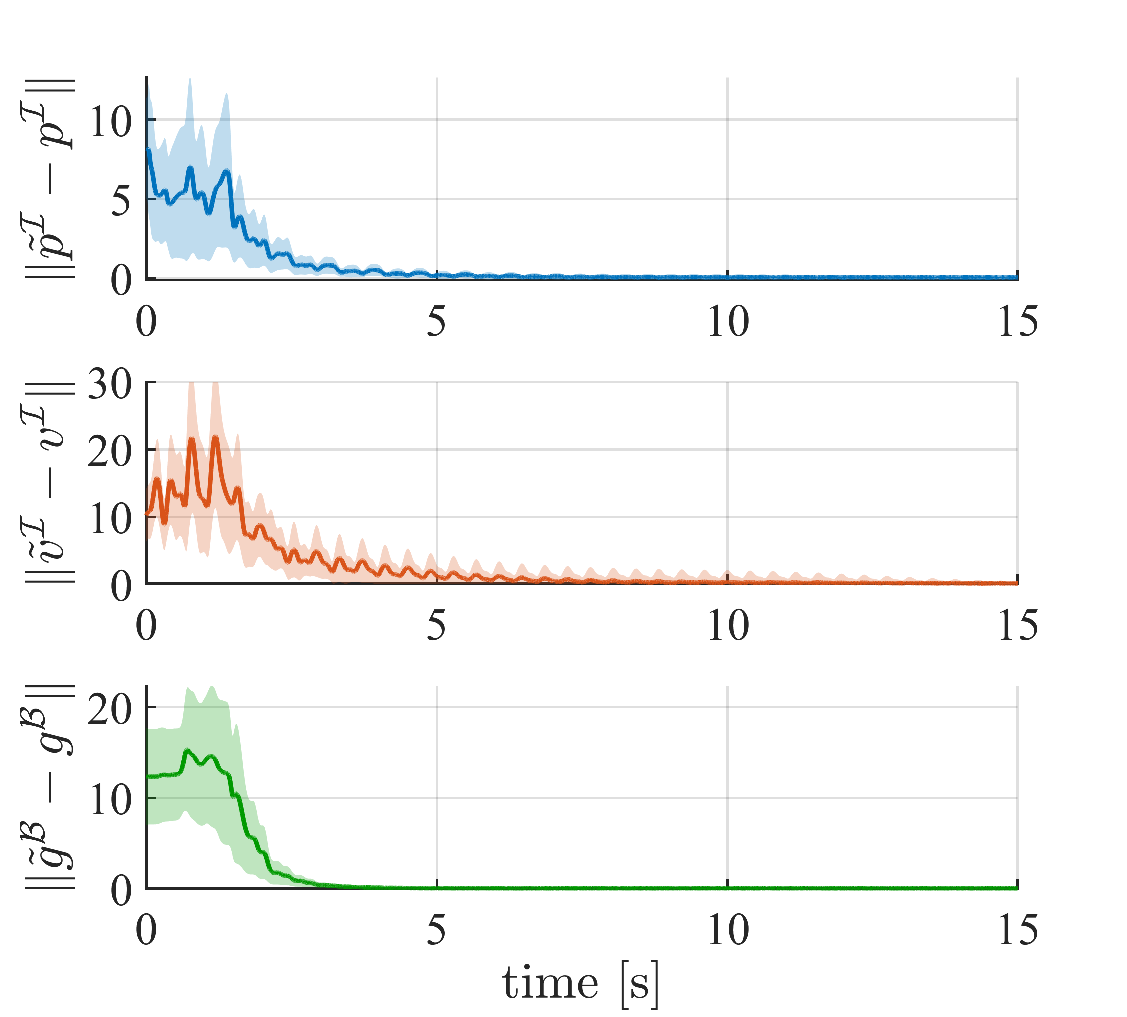}
    \caption{Position, velocity and gravity vector estimation error over time.}
    \label{fig:position_error}
\end{figure}

\begin{figure}%[h]
    \centering
    \includegraphics[width=\columnwidth]{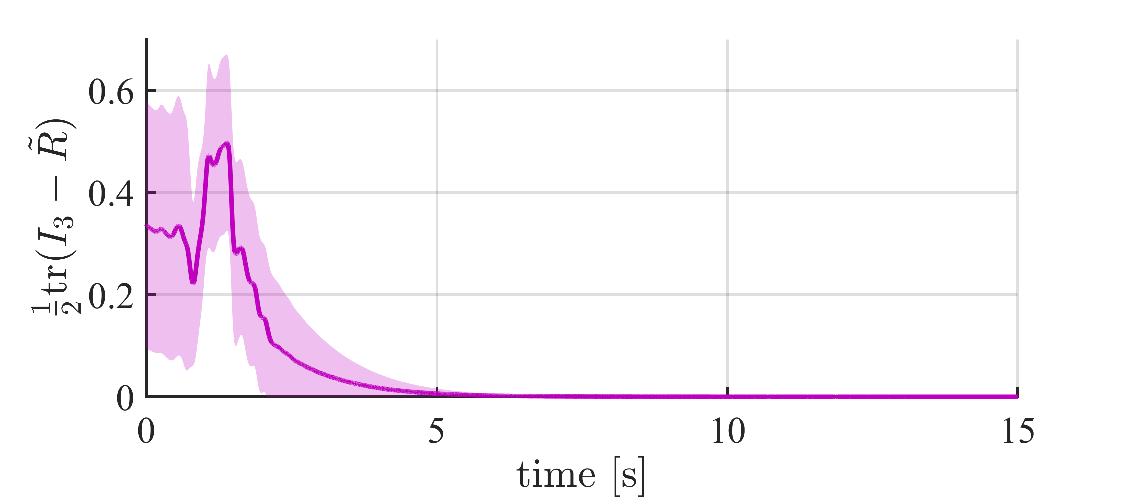}
    \caption{Orientation estimation error over time.}
    \label{fig:orientation_error}
\end{figure}

\begin{figure}%[h]
    \centering
    \includegraphics[width=0.7\columnwidth]{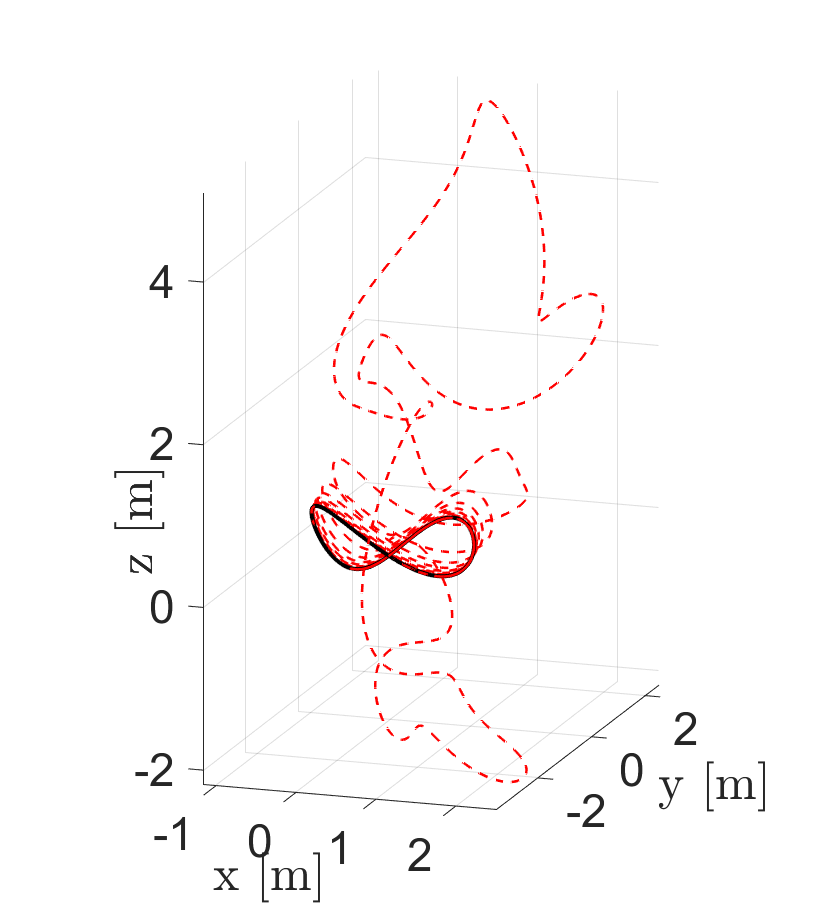}
    \caption{True trajectory of the vehicle in black line versus estimated trajectory in dashed red line}
    \label{fig:trajectory}
\end{figure}

\section{Conclusion}
In this paper, we proposed an observer for the full-state estimation of a rigid body freely navigating in three-dimensional space, using only a single range measurement combined with inertial and magnetometer data. By expressing the system dynamics in the body frame and augmenting the state vector with additional variables, we transformed the nonlinear estimation problem into a linear time-varying framework, suitable for  a Riccati observer design. We derived sufficient conditions ensuring uniform observability of the augmented system and demonstrated how these conditions could be obtained through a lower-dimensional system, significantly simplyfying the observability analysis. The proposed approach accurately estimates the position, velocity, and orientation, even under minimal sensor configurations and sensor noise. Simulation results confirmed the effectiveness and robustness of the observer. This approach introduces an efficient navigation solution, particularly suited for constrained environments where deploying multiple anchors is challenging. Future research directions include real life application, UWB clock offset estimation.

\section*{Appendix}
\subsection{Proof of Proposition \ref{prop1}}\label{appendixA}

We begin by noting that the uniform observability of the pair \((A(t), r_4)\) implies the uniform observability of the pair \((\mathcal{A}, \mathcal{C})\), which leads to the condition
\begin{align} \label{eq:gramian3}
    \int_t^{t+\delta_1} \Phi_{22}^\top(s, t) r_4^\top(s) r_4(s) \Phi_{22}(s, t) \, ds \geq \mu_1 I_{9}.
\end{align}

This result follows from Proposition~6 in \cite{Theodosis2021}. 
Since the augmented dynamics matrix $\mathcal{A}(t)$ is nilpotent and has an upper block-triangular structure, its state-transition matrix  $\Phi(t,s)$ inherits the same structure. Moreover, the output matrix $\mathcal{C}$ extracts only the last block of the  state, so the observability of the full augmented pair $(\mathcal{A}(t),\mathcal{C})$ depends exclusively on the dynamics of the lower subsystem governed by $A(t)$ and $r_4(t)$. Therefore, the uniform observability of $(\mathcal{A}(t),\mathcal{C})$ reduces to the uniform observability of the  reduced pair $(A(t), r_4(t))$.
The detailed proof is stated formally in proof of Proposition 6 of \citep{Theodosis2021}. Furthermore, due to the structure of the matrix \(A(t) = \bar{A} - \mathrm{blkdiag}([\omega]_{\times}, [\omega]_{\times}, [\omega]_{\times}\), it follows as shown in \citep{sifo2022} that the transition matrix \(\Phi_{22}(t, s)\) admits the following decomposition
\[
\phi_{22}(t, s) = \bar{R}^\top(t) \exp(\bar{A}(t - s)) \bar{R}(s) = \bar{R}^\top(t) \bar{\phi}(t, s) \bar{R}(s).
\]
Substituting this expression into the previous Gramian condition \eqref{eq:gramian3} yields:
\begin{multline}
\bar{R}(t)^\top \int_t^{t+\delta_1} 
\bar{\Phi}_{22}^{\top}(s, t) \bar{R}(t) r_4^{\top}(s) r_4(s) \\
 \bar{R}(t)^\top \bar{\Phi}_{22}(s, t) ds \bar{R}(t) \geq \mu_1 I_{9},
\end{multline}
thereby establishing the equivalence between the observability of the pair \((A(t), r_4)\) and that of the pair \((\bar{A}, \bar{R} r_4)\), which implies the equivalence between the observability of \((\mathcal{A}, \mathcal{C})\) and that of the pair \((\bar{A}, \bar{R} r_4)\).
\begin{flushright}
    $\square$
\end{flushright}

\subsection{Proof of Lemma \ref{TechnicalLemma}} \label{appendixB}
Let us  proceed by contradiction and Assume that the pair $(A,\Theta(t)H)$ is not uniformly observable. Then, for all $\bar\delta > 0$, there exits $x \in \mathbb{S}^{n-1}$ and a sequence of times $\{t_p\}_{p\in\mathbb{N}}$ such that
$$
\lim_{p\to\infty}\int_{0}^{\bar\delta} \|\Theta(t_p + s)H \exp(As)x\|^2 ds \ = \ 0
$$
which implies that
\begin{align} \label{contradiction}
   \lim_{p\to\infty}\int_{\bar\delta-\delta}^{\bar\delta} \|\Theta(t_p + s)H \exp(As)x\|^2 ds  =  0 
\end{align}
provided that $\bar\delta > \delta$. On the other hand, since $A$ is nilpotent the exponontial truncates is given as follows
$$
\exp(At) \ = \ I + tA + \cdots + \frac{t^{q-1}}{(q - 1)!} A^{q-1}.
$$
Multiplying by $H$
\begin{multline}
    H \exp(At)x \ = \ Hx + tHAx + \cdots + \frac{t^{q-1}}{(q - 1)!} HA^{q-1}x 
\\ =\ z_0 + t z_1 + \cdots + \frac{t^{q-1}}{(q - 1)!} z_{q-1}.
\end{multline}
Moreover, since $(A,H)$ is Kalman observable, not all $z_i$ are zero. Let us define 
$
\bar q \;:=\; \max\{\, i\in\{0,\dots,q-1\} \;\mid\; z_i\neq 0 \,\}.
$
which represent the leading nonzero index. With this choice we have  
\begin{align*}
    &\lim_{t\to\infty} \frac{H \exp(At)x}{t^{\bar q}} \ = \ z_{\bar q}\\
&\text{or, equivalently, } \qquad
\frac{H \exp(At)x}{t^{\bar q}} \ = \ z_{\bar q} + \eta(t), \ \ \eta(t)\to 0.
\end{align*}
where $\eta(t)$ captures precisely the contribution of the other coeifficient $z_0,\cdots, z_{\bar{q}-1}$. It is clear that $z_{\bar q}\in\mathbb{E}$. In fact if $z_{q-1}\neq 0$, we have $z_{\bar q}=z_{q-1}\in\textrm{Im}(HA^{q-1})\subseteq\mathbb{E}$. If $z_{q-1}=0$ and $z_{q-2}\neq0$ then $x\in\mathrm{Ker}(HA^{q-1})$ and hence $z_{\bar q}=z_{q-2}\in\textrm{Im}_{\mid L_{q-2}}(HA^{q-2})$ with $L_{q-2}=\mathrm{Ker}(HA^{q-1})$. The argument continues similarly. 
Now, let us pick $\delta$ large enough such that $\sup_{[\bar\delta-\delta,\bar\delta]} |\eta(s)| < \sqrt{\mu/4}\,\|z_{\bar q}\|/\bar\Theta$, where $\lVert \Theta(t) \rVert \leq \bar{\Theta}$. Then, using Young's inequality, one obtains
\begin{align*}
   \int_{\bar\delta-\delta}^{\bar\delta} &\|\Theta(t_p + s)H \exp(As)x\|^2 ds 
 \\
 &\ge (\bar\delta - \delta)^{\bar q} \!\!\int_{\bar\delta-\delta}^{\bar\delta} \|\Theta(t_p + s)(z_{\bar q} + \eta(s))\|^2 ds
\\&\ge\ (\bar\delta - \delta)^{\bar q} \!\Bigg(\frac{1}{2}\int_{\bar\delta-\delta}^{\bar\delta} \|\Theta(t_p + s)z_{\bar q}\|^2 ds  \\
&\hspace{1cm}-  \int_{\bar\delta-\delta}^{\bar\delta} \|\Theta(t_p + s)\eta(s)\|^2 ds \Bigg)\\
&\ge\ (\bar\delta - \delta)^{\bar q} \, (\mu\delta/2 - \mu\delta/4) \, \|z_{\bar q}\|^2 \ \\
&= \ (\bar\delta - \delta)^{\bar q} \, \mu\delta \, \|z_{\bar q}\|^2 / 4 \ > \ 0.
\end{align*}
This contradicts \eqref{contradiction}, and concludes the proof. \begin{flushright}
    $\square$
\end{flushright}

\subsection{Proof of Lemma \ref{lemma1}}\label{appendixC}
We will assume  that the signals 
$\ddot a^{\mathcal B}(t)$, $\dot a^{\mathcal B}(t)$, $a^{\mathcal B}(t)$, and $\omega(t)$ 
are bounded. 
Next, we verify that all the assumptions of Lemma~\ref{TechnicalLemma} hold for the reduced pair 
$(\bar A,\ \bar R(t) r_4(t))$. First, the pair $(\bar{A}, I_9)$ is Kalman observable. Second, $\bar A$ is constant and nilpotent, and finally, $\bar{R}r_4$ is bounded since all the signals involved are bounded. Then, we will use Lemma \ref{TechnicalLemma} to derive an explicit sufficient PE condition for the reduced pair $\big(\bar A,\ \bar R(t)r_4(t)\big)$. 

Let $ \int_{t}^{t+\delta_5}\!\!\|\bar R(s)\,r_4(s)\,z\|^2\,ds$, and for the constant matrix $\bar A\in\mathbb{R}^{9\times 9}$ one has
\[
\mathbb{E}=\ker(\bar A)=\Big\{\,z=\big[z_1^\top\ 0^\top\ 0^\top\big]^\top\in\mathbb{R}^9:\ z_1\in\mathbb{R}^3\Big\}.
\]
The fourth row $r_4(t)=[\,r_{4,1}(t)\;\;r_{4,2}(t)\;\;r_{4,3}(t)\,],\quad r_{4,1}(t)=\phi(t)^\top$ calculated using the recursive equation given in proposition \ref{prop1} is given by
$$
\phi(t):=\ddot a^{\mathcal B}(t)-[\omega(t)]_\times\dot a^{\mathcal B}(t)-2[\omega(t)]_\times^2 a^{\mathcal B}(t)\in\mathbb{R}^3,
$$
while $r_{4,2}(t)=4\,\dot a^{\mathcal B}(t)^\top-3\,a^{\mathcal B}(t)^\top[\omega(t)]_\times$ and $r_{4,3}(t)=6\,a^{\mathcal B}(t)^\top$. Hence, for any $z\in\mathbb{E}$ of the form $z=[z_1^\top\ 0^\top\ 0^\top]^\top$,
\[
r_4(t)z=r_{4,1}(t)z_1=\phi(t)^\top z_1,
 \text{and}
\|\bar R(t)\,r_4(t)z\|=\|r_4(t)z\|
\]
since $\bar R(t)=\mathrm{blkdiag}(R,R,R)$ is orthogonal. Therefore,
\begin{multline*}
    \int_{t}^{t+\delta_5}\!\!\|\bar R(s)\,r_4(s)\,z\|^2\,ds
=\int_{t}^{t+\delta_5}\!\!(\phi(s)^\top z_1)^2\,ds
\\=z_1^\top\!\left(\int_{t}^{t+\delta_5}\!\!\phi(s)\phi(s)^\top ds\right)\!z_1.
\end{multline*}
Since $ z \in \mathbb{E} \cap \mathbb{S}^8$, the norm $\lVert z_1 \rVert =1, \forall z_1 \in \mathbb{R}^3$  which result in the following PE condistion for the uniform stability of the system 
$$
\int_{t}^{t+\delta_5}\phi(s)\phi(s)^\top ds \geq \mu_5 I_3
$$
This is exactly the sufficient PE in Lemma~\ref{TechnicalLemma}, which ensures the uniform observability of the pair $\big(\bar A,\ \bar R(t)r_4(t)\big)$. Furthermore, since the observability of the full pair $(\mathcal{A}(t),\mathcal{C})$ 
is equivalent to the observability of the reduced pair $(\bar A,\ \bar R(t) r_4(t))$ 
(as established in Proposition~\ref{prop1}), the PE condition above guarantees the 
uniform observability of the augmented system~\eqref{aug_system}.
\begin{flushright}
    $\square$
\end{flushright}

\bibliography{ifacconf}   
\end{document}